\documentclass{styles/egpubl}
\usepackage{styles/egsr2025_DLOnlyTrack}
 
\WsPaper           %

\CGFccby

\usepackage[T1]{fontenc}
\usepackage{styles/dfadobe}  
\usepackage{relsize}

\biberVersion
\BibtexOrBiblatex
\usepackage[backend=biber,bibstyle=styles/EG,citestyle=alphabetic,backref=true]{biblatex} 
\electronicVersion
\PrintedOrElectronic

\ifpdf \usepackage[pdftex]{graphicx} \pdfcompresslevel=9
\else \usepackage[dvips]{graphicx} \fi

\usepackage{styles/egweblnk} 

\usepackage{multirow}
\usepackage{makecell}
\usepackage{colortbl}
\usepackage{xcolor}
\usepackage{xspace}
\usepackage{amsmath}
\usepackage{booktabs} 
\usepackage{amsfonts}
\usepackage{url}
\usepackage{bbold}
\usepackage[normalem]{ulem}
\usepackage{pifont}
\usepackage{nicefrac}

\newcommand{\citet}[1]{\textcite{#1}} %

\definecolor{rred}{RGB}{216, 27, 96}
\definecolor{rblue}{RGB}{30, 136, 229}
\definecolor{rgreen}{RGB}{28, 151, 77}
\definecolor{ryellow}{RGB}{255, 193, 7}
\definecolor{rorange}{RGB}{255,167,0}
\definecolor{americanrose}{rgb}{1.0, 0.01, 0.24}
\definecolor{cadmiumgreen}{rgb}{0.0, 0.42, 0.24}
\definecolor{ryb}{rgb}{0.1, 0.29, 0.8}
\definecolor{cc1}{HTML}{ee6352}
\definecolor{cc2}{HTML}{59cd90}
\definecolor{cc3}{HTML}{3fa7d6}
\definecolor{cc4}{HTML}{fac05e}
\definecolor{cc5}{HTML}{f7a9a8}
\definecolor{cc6}{HTML}{77b05f}
\definecolor{cc7}{HTML}{a64d79}
\definecolor{cc8}{HTML}{3c78d8}
\definecolor{c1}{HTML}{2196f3}
\definecolor{c2}{HTML}{4caf50}
\definecolor{c3}{HTML}{ff9800}
\definecolor{c4}{HTML}{f44336}
\definecolor{c5}{HTML}{673ab7}
\definecolor{originalcolor}{HTML}{D79B00}
\definecolor{transformedcolor}{HTML}{82B366}
\definecolor{author1}{HTML}{4D9DE0}
\definecolor{author2}{HTML}{CB48B7}
\definecolor{author3}{HTML}{33ab7c}
\definecolor{author4}{HTML}{1a7d19}
\definecolor{todo}{HTML}{E15554}

\newcommand{\edit}[1]{#1}

\newcommand{\method}{MatSwap}
\newcommand{\datasetname}{PBRand}
\newcommand{\nbscenes}{50,000}
\newcommand{\nbtotal}{250,000}
\newcommand{\nbvariants}{5}
\newcommand{\mask}{\ensuremath{\mathbf{M}}\xspace}
\newcommand{\normal}{\ensuremath{\mathbf{N}}\xspace}
\newcommand{\target}{\ensuremath{\mathbf{x}}\xspace}
\newcommand{\exemplar}{\ensuremath{\mathbf{p}}\xspace}

\newcommand{\xx}{\textsf{X}\xspace}

\newcommand{\rgbtoxx}{\textsf{RGB$\rightarrow$X}\xspace}
\newcommand{\xxtorgb}{\textsf{X$\rightarrow$RGB}\xspace}
\newcommand{\rgbtooxx}{\textsf{RGB$\leftrightarrow$X}\xspace}

\newcommand{\irra}{\ensuremath{\mathbf{E}}\xspace}
\newcommand{\image}{\ensuremath{\mathbf{I}}\xspace}

\newcommand{\best}[1]{\textbf{#1}}
\newcommand{\scnd}[1]{\underline{#1}}

\newcolumntype{C}[1]{>{\centering\let\newline\\\arraybackslash\hspace{0pt}}m{#1}}
\newcolumntype{S}[1]{@{\hspace{#1}}}
\newcolumntype{H}{>{\setbox0=\hbox\bgroup}c<{\egroup}@{}}

\newcommand{\verti}[2]{\multirow{1}{*}[#1]{\rotatebox[origin=c]{90}{#2}}}

\newcommand{\oc}[1]{\cellcolor{gray!10}}

\DeclareFontFamily{U}{mathb}{}
\DeclareFontShape{U}{mathb}{m}{n}{
  <-5.5> mathb5
  <5.5-6.5> mathb6
  <6.5-7.5> mathb7
  <7.5-8.5> mathb8
  <8.5-9.5> mathb9
  <9.5-11.5> mathb10
  <11.5-> mathb12
}{}
\DeclareSymbolFont{mathb}{U}{mathb}{m}{n}
\DeclareMathSymbol{\ulsh}{3}{mathb}{"E8}
\DeclareMathSymbol{\ursh}{3}{mathb}{"E9}
\DeclareMathSymbol{\dlsh}{3}{mathb}{"EA}
\DeclareMathSymbol{\drsh}{3}{mathb}{"EB}

\makeatletter
\DeclareRobustCommand\onedot{\futurelet\@let@token\@onedot}
\def\@onedot{\ifx\@let@token.\else.\null\fi\xspace}

\def\eg{\emph{e.g}\onedot} 
\def\ie{\emph{i.e}\onedot} 
\def\cf{\emph{cf}\onedot}

\makeatother

\usepackage[capitalize]{cleveref}
\crefname{section}{sec.}{secs.}
\Crefname{section}{Sec.}{Secs.}
\crefname{paragraph}{sec.}{secs.}
\Crefname{paragraph}{Sec.}{Secs.}
\crefname{table}{Tab.}{Tabs.}
\Crefname{table}{Tab.}{Tabs.}
\crefname{figure}{Fig.}{Figs.}
\Crefname{figure}{Fig.}{Figs.}
\crefname{equation}{eq.}{eqs.}
\Crefname{equation}{Eq.}{Eqs.}

\title[\method{}: Light-aware material transfers in images]%
      {\method{}: Light-aware material transfers in images}

\author[I. Lopes \& V. Deschaintre \& Y. Hold-Geoffroy \& R. de Charette]
{\parbox{\textwidth}{\centering I. Lopes$^{1}$\orcid{0009-0001-3755-7529} \quad V. Deschaintre$^{2}$\orcid{0000-0002-6219-3747} \quad Y. Hold-Geoffroy$^{2}$\orcid{0000-0002-1060-6941}  \quad R. de Charette$^{1}$\orcid{0000-0003-3738-1962} \\[1ex] \centering $^1$Inria \, $^2$Adobe Research}}

\begin{document}

\teaser{
    \includegraphics[width=\textwidth]{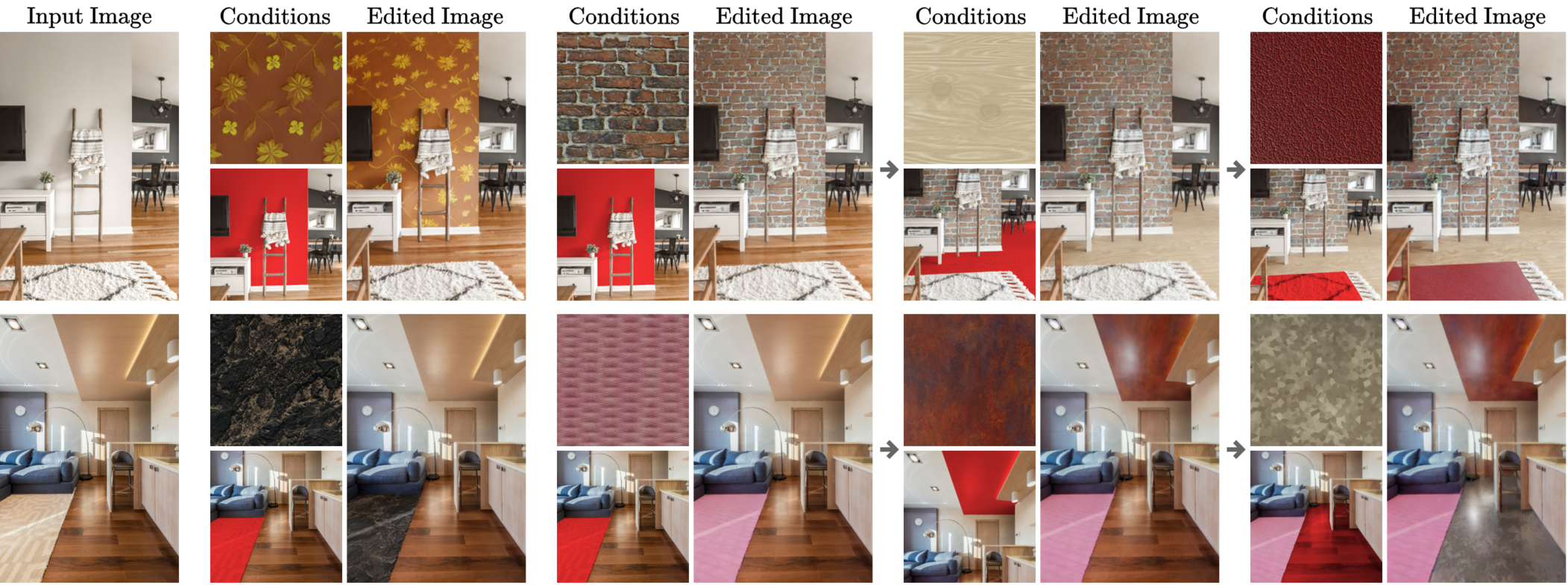}
    \caption{\textbf{\method{} allows realistic material transfer in images.} From an input image (left), our method seamlessly integrates an exemplar material (top left inset) into the user-specified region (red mask, bottom left inset). We can plausibly replace the wall's surface (top) with tapestry (first result) or bricks (second), and also alter the wood type on the floor (third) and the mat (rightmost). Similarly, we present two distinct material swaps for the carpet (bottom row), and further modify our second result by changing the ceiling and floor (rightmosts). \method{} realistically handles lighting effects such as low-frequency shading (wall, top row) and cast lights (mat, bottom row). }
    \label{fig:teaser}
}

\maketitle
\begin{abstract}
We present~\method{}, a method to transfer materials to designated surfaces in an image realistically.
Such a task is non-trivial due to the large entanglement of material appearance, geometry, and lighting in a photograph. In the literature, material editing methods typically rely on either cumbersome text engineering or extensive manual annotations requiring artist knowledge and 3D scene properties that are impractical to obtain. 
In contrast, we propose to directly learn the relationship between the input material---as observed on a flat surface---and its appearance within the scene, without the need for explicit UV mapping. To achieve this, we rely on a custom light- and geometry-aware diffusion model. We fine-tune a large-scale pre-trained text-to-image model for material transfer using our synthetic dataset, preserving its strong priors to ensure effective generalization to real images. As a result, our method seamlessly integrates a desired material into the target location in the photograph while retaining the identity of the scene. \method{}~is evaluated on synthetic and real images showing that it compares favorably to recent works. Our code and data are made publicly available on \url{ https://github.com/astra-vision/MatSwap} 

\ccsdesc[500]{Computing methodologies~Texturing}
\ccsdesc[500]{Computing methodologies~Image processing}
\ccsdesc[300]{Computing methodologies~Rendering}
\ccsdesc[100]{Computing methodologies~Reflectance modeling}
\ccsdesc[100]{Computing methodologies~Computer vision}

\printccsdesc   
\end{abstract}  

\section{Introduction}
\vspace{-2mm}
Photographs capture the visual appearance of a scene by measuring the radiant energy resulting from the complex interaction of light, geometry, and materials. Among others, textures and materials are key components that contribute to the aesthetics and emotions conveyed by images~\cite{joshi2011aesthetics}. %
Unfortunately, their appearances are largely entangled with the scene's lighting and geometry, making it difficult to edit a posteriori. \\
Recently, the editing or generation of images has been significantly simplified by advancements in diffusion models that can benefit from internet-scale datasets~\cite{schuhmann2022laionb}. Such models can be used for prompt-guided diffusion inpainting, where only part of an image is modified to follow user guidance~\cite{meng2022sdedit,lugmayr2022repaint}. However, textually describing a material is not trivial, especially when it {exhibits} complex patterns or appearance. An alternative is to use pixel-aligned maps to drive the generation. 
Using ControlNet~\cite{zhang2023controlnet} such conditions can come in the form of semantics (\eg, segmentation maps), visual maps (\eg, edges), or geometry information (\eg, depth, normal). 
IP-Adapter~\cite{ye2023ipadapter} proposes a similar approach for global conditioning, where CLIP~\cite{radford2021learning} visual embeddings are used as an effective guidance signal for image generation. 
ZeST~\cite{cheng2024zest} builds on the latter, showing that material inpainting can be conditioned with a CLIP encoder to extract a material appearance from an image. This idea was recently further developed by Garifullin et al.~\cite{garifullin2025materialfusion}, where the CLIP conditioning is enhanced with self-guidance to help preserve content identity. However, these approaches offer little control to the artist over the transferred material appearance (\eg, scale, rotation). 
To achieve greater control, \rgbtooxx~\cite{zeng2024rgbx} proposes to directly modify the estimated PBR maps, at the cost of manual per-pixel editing. However, such an approach is impractical for spatially varying materials, as manual editing requires careful geometry and perspective texture projection handling.

In this work, we introduce \method, an exemplar-based method that improves material transfer in images while simplifying its usage and offering more controllability to the user. 
Given an image and a texture sample rendered or photographed on a mostly flat surface, our method inpaints a region of the image using the texture sample (\cf~\cref{fig:teaser}), ensuring a close alignment with the original geometric and illumination cues obtained from off-the-shelf single image estimators~\cite{zeng2024rgbx,he2024lotus}. 

More precisely, we rely on a light- and geometry-aware diffusion model training for generating realistic inpainted images. We fine-tune a pre-trained diffusion model~\cite{rombach2021highresolution}, which already incorporates priors about object appearances in images, using a new synthetic dataset. Our goal is therefore to specialize our model to the material transfer task while retaining its original priors for better generalization to real photographs. 
To train our model, we generate a procedural 3D dataset named \datasetname{}, which consists of primitive shapes randomly arranged and lit with captured environment maps, and render pairs of images with varying materials applied on a randomly selected object surface.
Finally, we compare \method{} to state-of-the-art inpainting~\cite{rombach2021highresolution,podell2023sdxl,flux.1,avrahami2023blended} and the latest material transfer methods~\cite{cheng2024zest,zeng2024rgbx,garifullin2025materialfusion} showing that it outperforms them qualitatively and quantitatively. 

Our approach enables material replacement in images through the following contributions:
\begin{itemize}
    \item A light- and geometry-aware diffusion model that performs material transfer in a single image;
    \item A two-stage training pipeline based on synthetic data that generalizes to real images;
    \item \datasetname{}, a new synthetic procedural dataset of primitive scenes that provides \nbtotal~paired renderings suitable for training material replacement methods.
\end{itemize}

\begin{figure*}[t]
  \includegraphics[trim={0 0 1.5cm 0},clip,width=.98\textwidth]{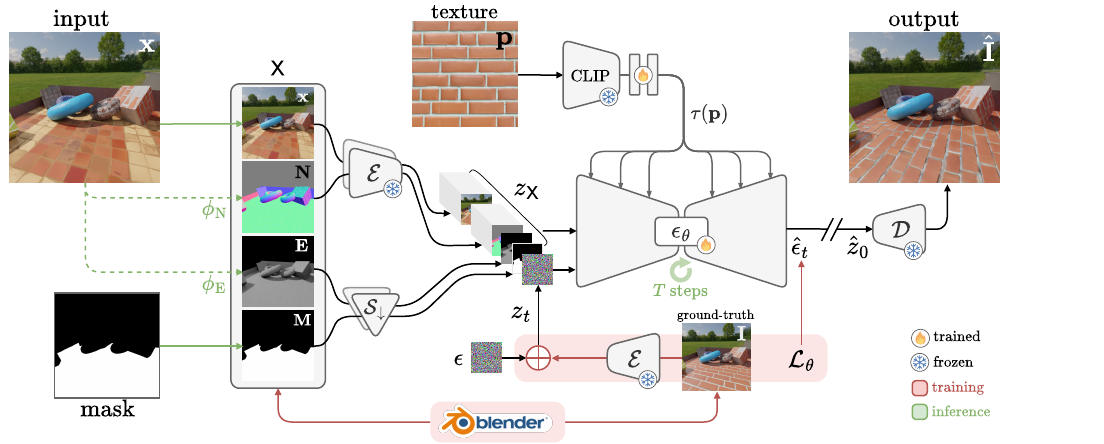}
  \caption{\textbf{Overview of \method{}.} We learn to transfer the texture \exemplar on a given region \mask of an input image \target by training a light- and geometry-aware diffusion model, leveraging irradiance \irra and normal \normal maps. Once encoded ($\mathcal{E}$) or downsampled ($\mathcal{S}_\downarrow$), the image, mask, and maps are concatenated into a scene descriptor $z_X$ which, together with the noise latent $z_t$, serve as input to the denoising UNet, $\epsilon_{\theta}$. To integrate the exemplar conditioning, we inject the visual CLIP features of the texture via adapter layers from IP-Adapter~\cite{ye2023ipadapter}. 
  During training all scene descriptors are obtained from Blender (bottom), while at inference, only \target{} and the mask \mask are accessible so we leverage off-the-shelf estimators ($\phi_N$, $\phi_E$) to extract normals \normal and irradiance \irra from the input image (left).
  }
  \label{fig:method}
\end{figure*}

\section{Related works}

\subsection{Image generation and editing}
Neural image generation received significant attention over the past decade, starting with GANs~\cite{goodfellow2014gans,Karras2019stylegan2}. Lately, diffusion models have become the \textit{de facto} image generation framework~\cite{sohl2015deep,rombach2021highresolution,podell2023sdxl,flux.1,peebles2023dit,betker2023dalle3}%
, producing high-quality results alongside normalizing flows~\cite{zhang2021diffusionflow,esser2024sd3} and benefiting from internet-scale datasets~\cite{schuhmann2022laionb}.

Controlling these image generation models became an active research field, first using text~\cite{ho2021classifier,meng2022sdedit}, then using various image modalities~\cite{zhang2023controlnet} or physically-based rendering (PBR) properties~\cite{zeng2024rgbx}. An effective approach to conditioning diffusion models is to train or fine-tune them with the desired control as input, for instance, using an image~\cite{ke2023marigold,he2024lotus}, or other scene properties~\cite{zeng2024rgbx}. 
An alternative is to inject the conditioning maps into the pre-trained frozen diffusion model, either using a parallel network~\cite{zhang2023controlnet}, an adapter \cite{ye2023ipadapter}, or via low-rank adaptation of the text encoder \cite{lopes2024material}. 
Another control is to directly manipulate the input image embedding to modify its semantic properties~\cite{guerrero2024texsliders}. Finally, operations on attention maps~\cite{hertz2022prompt,epstein2023diffusion,parmar2023zero} are also commonly used to manipulate parts of images during the generation process. 

These conditionings can be coupled with the task of inpainting, when part of an image is generated to be seamlessly integrated into the input image. Generative diffusion models can be used for inpainting by compositing, at every denoising step, the estimated latent of the inpainted region with the latent of the input image \cite{lugmayr2022repaint}. However, this leads to visible artifacts along the mask boundaries \cite{cheng2024zest}. 
Text-driven local image editing was proposed in Blended Latent Diffusion \cite{avrahami2023blended}, blending latents using a mask during the denoising process.

In this work, we train an existing diffusion model~\cite{rombach2021highresolution} to perform material replacement. We incorporate additional inputs through zero-weight addition~\cite{zhang2023controlnet}. 
To avoid the complexities of UV mappings linked to pixel-aligned conditionings, we leverage the priors of the diffusion model regarding object appearance, paired with global conditioning as suggested by IP-adapter~\cite{ye2023ipadapter}, to define the desired material appearance.

\subsection{Environment-aware image editing}
Changing the appearance of a surface within an image is trivial in a 3D editor, but proves very challenging in photographs due to the complex interactions conflating appearance, light, and geometry.

\textbf{Lighting} plays a crucial role in photorealism and is a clear sign of forgery when not correctly handled \cite{kee2013exposing}. When editing images, one way to encode lighting is through radiance, either using a parametric \cite{griffiths2022outcast,gardner2019deep,poirier2024diffusion} or non-parametric light model \cite{pandey2021total,gardner2017learning}. However, radiance is challenging for deep learning models due to its high dynamic range and spherical nature, making it difficult to map to the image plane. Inspired by intrinsic image decomposition, recent image editing methods chose irradiance as encoded by shading maps to represent illumination and perform object insertion \cite{zhang2024zerocomp,fortier2024spotlight} or relighting \cite{kocsis2024lightit,ponglertnapakorn2023difareli,yu2020self}. Our method uses this same irradiance representation, estimating shading maps using \rgbtooxx~\cite{zeng2024rgbx}. 

\textbf{Material} replacement is a long-standing problem in Computer Graphics~\cite{an2008appprop, khan2006image} with early methods proposing to adjust materials reflectance of color~\cite{an2008appprop} through user scribble and edit propagation, or changing materials to metallic or glossy~\cite{khan2006image}. Leveraging deep learning, methods were proposed to edit materials in photographs, often targeting textures~\cite{guerrero2024texsliders} or objects~\cite{delanoy2022generative,cheng2024zest,sharma2024alchemist}. On a scene scale, \rgbtooxx~\cite{zeng2024rgbx} proposed using PBR maps as control, enabling the per-pixel change of material properties, in particular the albedo. However, this requires manual, per-pixel editing, which is impractical for textures or perspective-distorted objects. 
Closest to our work is ZeST~\cite{cheng2024zest}, which proposes a training-free method based on IP-Adapter to perform material transfer. This concept is expanded by MaterialFusion~\cite{garifullin2025materialfusion} by adding self-guidance \cite{epstein2023diffusion} to ZeST's conditioning to help preserve details and identity from the original image.
Our method differs from ZeST and MaterialFusion in both formulation and implementation. Instead of treating material transfer as an inpainting task, we generate the entire image, which enhances context during diffusion and minimizes artifacts (see zoom-in in \cref{fig:baselines}). %
Further, we leverage off-the-shelf estimators for more accurate guidance (\eg, illumination). These differences lead to improved transfers, more accurate shading and fewer artifacts around the mask edges.

\section{Material Transfer}

Material transfer involves applying a material to a designated surface in an image, ensuring it integrates seamlessly into the scene.
It can be seen as a form of 3D-aware inpainting, where a given material is plausibly blended within a target image while preserving its shading and geometry cues. 

Specifically, given an exemplar texture image~\exemplar, a target image~\target, and a target mask~\mask, our material transfer consists of replacing the region defined by \mask in \target with a material that resembles the exemplar texture image~\exemplar. 
The mask is arbitrarily defined by the user and can cover either an object or a surface, or be obtained by an automatic segmentation method \cite{kirillov2023segment,sharma2023materialistic}. None of the inputs require time-consuming annotations or expertise in 3D modeling, such as UV maps or texture wrapping. 

To accomplish this, we build upon a latent diffusion model that encodes an image \image into a latent space represented by an encoder: $z_0 = \mathcal{E}(\image)$. From this point, we carry out iterative denoising as outlined in \cite{ho2020denoising}. 
We start with a Gaussian-sampled latent vector $z_T$ and aim to produce its denoised counterpart, the latent vector $z_0$. This is a multi-step process, where a UNet predicts the residual $\epsilon$ to denoise the latent variable at every step.

\subsection{\method}
\label{sec:meth_meth}
We consider material transfer as a conditional generative task leveraging diffusion models. 
While the literature typically relies solely on geometrical cues~\cite{zhang2023controlnet,cheng2024zest}, we observe that this can lead to inconsistent shading. Therefore, we guide our diffusion process with both scene illumination and geometry. 

Assuming an image \target of a scene, we describe it with a conditioning \xx which includes the target image and mask along with pixel-wise intrinsic maps representing normals \normal %
and diffuse irradiance \irra. 
Importantly, we ensure that these maps do not contain material information, avoiding to include properties such as albedo, roughness, or metalness. This ensures the diffusion model is provided with enough information about the scene structure and illumination while remaining material-independent. 
Subsequently, as \irra we choose to represent only the diffuse illumination of the scene without specular effects since these are too correlated with materials, making it impractical to condition the model. 
We train our method on synthetic data (\cref{sec:meth_dataset}), using the conditioning buffers readily provided by rendering engines while demonstrating that our model successfully generalizes to real images. 
For the latter, we leverage recent advances in single-image intrinsic channel estimation to obtain reasonably accurate maps from off-the-shelf methods such as \rgbtooxx~\cite{zeng2024rgbx} or Lotus~\cite{he2024lotus}. Thus, for real images we define $\irra=\phi_\irra(\target)$ and $ \normal=\phi_\normal(\target)$, with $\phi_\normal$ and $\phi_\irra$ the normal and an irradiance estimators, respectively.

Next, we explain how to integrate the exemplar texture conditioning into the framework. 
Recently, IP-Adapter~\cite{ye2023ipadapter} demonstrated that image-prompt guidance could be accomplished by training adapters between the CLIP \cite{radford2021learning} visual encoder and the denoising UNet. Further works~\cite{vecchio2024controlmat,cheng2024zest,Yan:2023:PSDR-Room,guerrero2024texsliders} have shown that CLIP can be used to extract rich material features from images. 
Similarly, we condition our pipeline on the CLIP image embedding of the material we want to transfer in the target image. We replace the standard text cross-attention mechanism, injecting the visual CLIP features via adapter layers instead.

Our proposed method is illustrated in \cref{fig:method}. We encode the ground-truth image \image{} as $z_0 = \mathcal{E}(\image)$ and its scene descriptor stack $\xx=\{\target, \normal, \irra, \mask\}$ describing our target image \target, defined as:
\begin{equation}
    z_\xx = \left( \mathcal{E}(\target),\, \mathcal{E}(\normal),\, \mathcal{S}_\downarrow(\irra),\, \mathcal{S}_\downarrow(\mask) \right),
    \label{eq:encoding}
\end{equation} 
where $\mathcal{E}$ is a pre-trained latent encoder and $\mathcal{S}_\downarrow$ is a down-sampling operator. As seen in~\cref{eq:encoding}, both the target image~$\target$ and normal maps~$\normal$ are encoded while the irradiance map~$\irra$ and the inpainting mask~$\mask$ are downsampled following previous works~\cite{zeng2024rgbx,rombach2021highresolution}.
To provide the conditioning signal, $z_\xx$ is concatenated to the noisy input latent $z_t$ at every timestep $t$. The diffusion loss is defined as:
\begin{equation}
    \mathcal{L}_{\theta} = \left\| \epsilon_t - \epsilon_{\theta} \left( z_t, z_\xx, t, \tau(\exemplar) \right) \right\|^2_2.
\end{equation}
\noindent We write the denoising UNet $\epsilon_{\theta}$, with parameters $\theta$. It receives two kinds of inputs: the noisy image $z_t$ concatenated with the scene descriptor $z_\xx$; and a global conditioning via the cross-attention layers of $\epsilon_{\theta}$ containing the timestep $t$ and a CLIP embedding $\tau(\exemplar)$, with \exemplar being the exemplar texture image. To condition the diffusion process with visual features, we initialize the adapter weights with those from IP-Adapter \cite{ye2023ipadapter} and freeze the image encoder.
In our pipeline, we train the full UNet and drop the text prompt to rely exclusively on the image-prompt embedding $\tau(\exemplar)$, for which we train adapter layers, as seen in \cref{fig:method} \textit{top}.

Our end-to-end training method uses modality dropout on the conditioning latents of $z_\xx$ \cite{zeng2024rgbx} by randomly setting these to null vectors. 
This ensures the model can inpaint the target region with partial conditioning or even completely unconditionally. 
We also randomly drop the exemplar texture image to leverage classifier-free guidance (CFG) \cite{ho2021classifier}. This method is commonly used in text-to-image diffusion models to strengthen the input text conditioning. Here, we adopt this mechanism in the context of texture-conditioned diffusion. In practice, it corresponds to sampling $\epsilon_\theta$, once conditionally $\hat{\epsilon}_{\exemplar} = \epsilon_{\theta} \left( z_t, z_\xx, t, \tau(\exemplar) \right)$ and unconditionally $\hat{\epsilon}_{\emptyset} = \epsilon_{\theta} \left( z_t, z_\xx, t, \emptyset \right)$. At each timestep, $\hat{\epsilon}$ is obtained as: 
\begin{equation}
    \hat{\epsilon} = \hat{\epsilon}_{\emptyset} + \gamma \left( \hat{\epsilon}_{\exemplar} - \hat{\epsilon}_{\emptyset} \right)
\end{equation}
Here, $\gamma \geq 1$ represents the guidance scale. When $\gamma = 0$, the sampling is entirely unconditional, while the default conditional sampling occurs when $\gamma = 1$. We find that integrating CFG leads to significant improvement in transfer quality as shown in~\cref{fig:cfg}. 

\subsection{Dataset}
\label{sec:meth_dataset}

\begin{figure}[h!]
    \centering
    \setlength{\tabcolsep}{.5pt}
    \renewcommand{\arraystretch}{0.1}
    \newcommand{\imgsize}{.19\linewidth}

    \newcommand{\row}[2]{%
        \verti{10ex}{envmap #1} & \includegraphics[width=\imgsize,height=\imgsize]{images/dataset/#2_irradiance.png} & \includegraphics[width=\imgsize,height=\imgsize]{images/dataset/#2_normal.png} & \includegraphics[width=\imgsize,height=\imgsize]{images/dataset/#2_target.png} & \includegraphics[width=\imgsize,height=\imgsize]{images/dataset/#2_mask.png} & \includegraphics[width=\imgsize,height=\imgsize]{images/dataset/#2_texture.png} & \includegraphics[width=\imgsize,height=\imgsize]{images/dataset/#2_image.png}
    }
    
	\resizebox{.99\linewidth}{!}{%
    \begin{tabular}{c cccccc}
        & irradiance \irra & normal \normal & target \target & mask \mask & texture \exemplar & image \image \\[.5ex]
        \row{1}{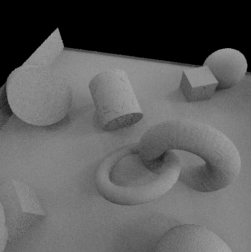} \\
        \verti{1.5ex}{...} \\
        \row{N}{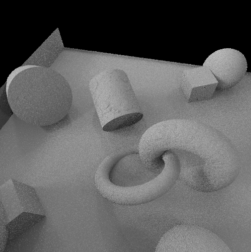} \\[.5ex]
        \row{1}{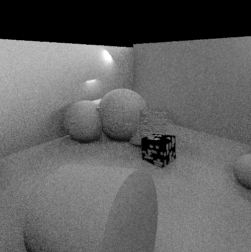} \\
        \verti{1.5ex}{...} \\
        \row{N}{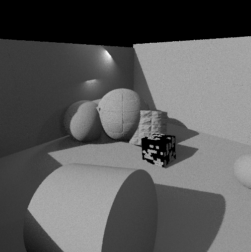} \\
    \end{tabular}}
    
    \caption{\textbf{Procedural dataset.} We show examples of our \datasetname{} dataset, which we use for training. It consists of primitive objects (spheres, cubes, cylinders, and tori) with random placements, orientations, and materials enclosed within four walls of varying heights. A total of \nbscenes{} 3D scenes were created in Blender, each rendered under $\nbvariants{}$ light variations, with image-based lighting to achieve realistic occlusions and cast shadows. 
    For every scene, we render a second scene identical to the first, except for one object for which we swap the material. Under ``texture \exemplar'', we show the full texture as well as a crop (outlined in white) that has a matching scale with the rendered surface.}
    \label{fig:dataset}
\end{figure}

To train our method, we need paired images showing the same scene with identical lighting but with a known material change.
We design a simple procedural 3D dataset in Blender, named \datasetname, consisting of \nbtotal~scene pairs which we render along with irradiance, normals, UV and material segmentation maps. Scenes are created by randomly placing 3D primitives and lights within the boundaries of a cubic scene, and randomly varying the wall heights to allow direct lighting and occlusions. The images are rendered with image-based lighting \cite{debevec2008rendering} using a randomly rotated environment map sampled from a set of 100 HDRIs~\cite{polyhaven}. We use approximately 4,000 unique materials from MatSynth~\cite{vecchio2023matsynth}, randomly assigned to all objects. For each scene, one of the objects is randomly selected, and its material is replaced with another. This generates two buffers per scene, each containing $\left( \target, \normal, \irra, \mask, \exemplar, \image \right)$ with only the material on the selected object changed. We show samples from our dataset in~\cref{fig:dataset}. 

The normal maps \normal{} in \datasetname{} include both scene geometry and material surface normals, similar to maps produced by standard normal estimators in existing literature \cite{zeng2024rgbx, he2024lotus,kocsis2024intrinsic}. We use these normal maps \normal{} directly as conditioning during inference, denoted as $\phi_{\normal}$. This lets our model learn how to best integrate the texture provided as input with the geometry normals of the scene during inference.

To enforce consistency between the scale of the conditioning image and the rendered texture, we measure the texture coverage from the scene UV buffer and scale the exemplar image \exemplar accordingly, similarly to recent work~\cite{ma2024materialpicker}. This ensures that the rendered texture appears at the same scale as in the conditioning image~(\eg,~a brick texture will have roughly the same number of tiles in~\exemplar and~\image). 
Despite the simplicity of~\datasetname, we found it sufficient for our model to learn strong priors for material transfer.

\subsection{Implementation Details}

Our model is based on Stable Diffusion~\cite{rombach2021highresolution}, a large publicly available text-to-image model. \edit{We use the adapter layers of IP-Adapter~\cite{ye2023ipadapter}, injecting information at 16 cross-attentions throughout the UNet. We concatenate additional input channels to the first convolution of $\epsilon_{\theta}$ resulting in 14 channels in latent space -- \normal: 4D, \irra: 1D, \mask: 1D, \target: 4D, $z_t$: 4D. These new channels are initialized with zero weights to prevent disrupting the training during its early stages.}
We train in two phases: first at a resolution of $256^2\mathrm{px}$ for 100k iterations, followed by 50k iterations at $512^2\mathrm{px}$ resolution. The training employs a batch size of 64 and spans roughly five days on a single Nvidia A100 GPU. The AdamW optimizer \cite{kingma2014adam,loshchilov2017fixing} is used with a learning rate of $2\!\times{}\!10^{-5}$.

To enhance robustness, we apply horizontal flipping as data augmentation, taking care to adjust the normals accordingly. Additionally, we implement a $10\%$ probability of dropping all the conditioning inputs except the mask, along with another $10\%$ chance of dropping any of the inputs: \irra, \normal, \target, \exemplar. To drop \exemplar{}, we set the CLIP embedding to the null vector. The mask \mask is always kept. 

Our input \exemplar corresponds to a material rendered on a flat surface covering the whole image in a fronto-parallel setting, illuminated with a random HDRI from PolyHaven~\cite{polyhaven}. During training, we utilize the materials from MatSynth~\cite{vecchio2023matsynth}, from which we extract 16 random crops. We account for the texture scale {in} the reference image~\target{} by extracting the UV coordinates of each region within the image, and cropping all samples~\exemplar{} by a factor of $\left( \max{\left( \mathrm{UV} \right)} - \min{\left( \mathrm{UV} \right)}\right)^{-1}$ in both horizontal and vertical dimensions. These exemplars are then resized to $224\!\times{}\!224\;\mathrm{px}$ to be fed as input to the CLIP encoder. This step is only performed during training, with the UV buffers provided by the rendering engine. We do not need UV mappings during inference when editing images, as the scale is set relatively, as shown in \cref{fig:scale}.

\begin{figure}
    \centering
    \setlength{\tabcolsep}{0pt}
    \renewcommand{\arraystretch}{0}
    \newcommand{\imgsize}{.2\linewidth}
	
    \resizebox{.99\linewidth}{!}{%
    \begin{tabular}{ccccc}
        \includegraphics[width=\imgsize]{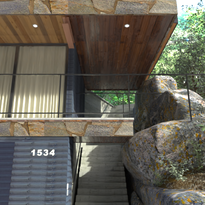}&%
        \includegraphics[width=\imgsize]{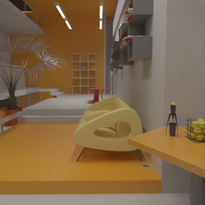}&%
        \includegraphics[width=\imgsize]{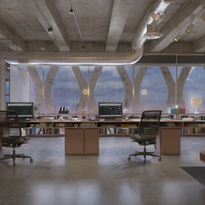}&%
        \includegraphics[width=\imgsize]{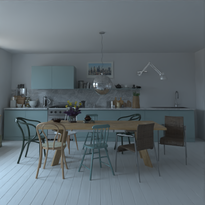}&%
        \includegraphics[width=\imgsize]{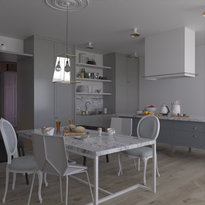}\\ 
        
        \includegraphics[width=\imgsize]{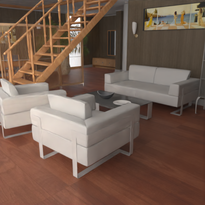}&%
        \includegraphics[width=\imgsize]{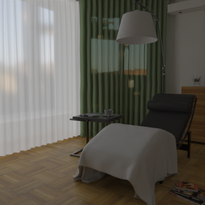}&%
        \includegraphics[width=\imgsize]{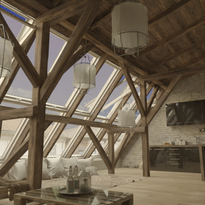}&%
        \includegraphics[width=\imgsize]{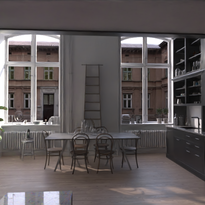}&%
        \includegraphics[width=\imgsize]{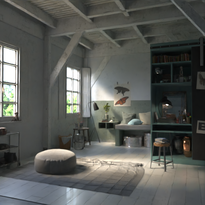}\\ 
    \end{tabular}}
    \caption{\textbf{Samples from synthetic evaluation dataset.} Scenes show a wide diversity in appearance and illumination.}
    \label{fig:archviz}
\end{figure}

\begin{figure*}
    \centering
    \setlength{\tabcolsep}{0pt}
    \renewcommand{\arraystretch}{0}
    \newcommand{\imgsize}{.125\textwidth}
    \newcommand{\imgsizehalf}{.0625\textwidth}
    \newcommand{\shift}{7.7ex}
    \newcommand{\prompt}{\cellcolor{gray!40}}
    
    \newcommand{\img}[1]{%
        \multirow{2}{*}[\shift]{\includegraphics[width=\imgsize,height=\imgsize]{#1}}%
    }
    \newcommand{\row}[4]{%
        &&\includegraphics[width=\imgsizehalf,height=\imgsizehalf]{images/baselines/texture_#1.png}%
        & \multirow{2}{*}[\shift]{\includegraphics[width=\imgsize,height=\imgsize]{images/baselines/bld_#1.png}}%
        & \multirow{2}{*}[\shift]{\includegraphics[width=\imgsize,height=\imgsize]{images/baselines/flux_#1.png}}%
        && \multirow{2}{*}[\shift]{\includegraphics[width=\imgsize,height=\imgsize]{images/baselines/normalscontrolnet_#1.png}}%
        & \multirow{2}{*}[\shift]{\includegraphics[width=\imgsize,height=\imgsize]{images/baselines/rgbx_#1.png}}%
        & \multirow{2}{*}[\shift]{\includegraphics[width=\imgsize,height=\imgsize]{images/baselines/zest_#1.png}}%
        & \multirow{2}{*}[\shift]{\includegraphics[width=\imgsize,height=\imgsize]{images/baselines/matfusion_#1.png}}%
        & \multirow{2}{*}[\shift]{\includegraphics[width=\imgsize,height=\imgsize]{images/baselines/ours_#1.png}} \\
        
        \multirow{2}{*}[12ex]{\verti{0ex}{\edit{row #4}}}%
        &\multirow{2}{*}[15.6ex]{\includegraphics[width=\imgsize,height=\imgsize]{images/baselines/image_#1.png}} %
        & \includegraphics[width=\imgsizehalf,height=\imgsizehalf]{images/baselines/mask_#1.png}  \\
        
    }

    \newcommand{\rowsub}[1]{
    }
    
	\resizebox{.99\linewidth}{!}{%
    \begin{tabular}{C{2ex} cS{.2ex}c S{.2ex}S{.2ex} c S{.2ex} c cS{.2ex} c S{.2ex} c S{.2ex} c S{.2ex} c S{.2ex} c}
        && \multicolumn{1}{c}{} & \multicolumn{2}{c}{\small \prompt{} \texttt{Text-prompt based}} && \multicolumn{5}{c}{\small \prompt{} \texttt{Image-prompt based}} \\[0.2ex]
        \\[0.5ex]
        &\multicolumn{1}{c}{Image} & Source & Blended LD & FLUX.1 Fill && SD2 Ctrl-\normal & \rgbtooxx & ZeST & Material Fusion & \textbf{ours} \\[.7ex]
        
        \row{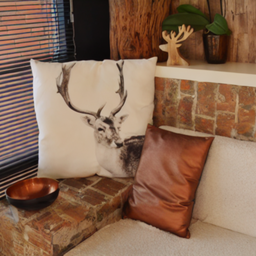}{Aged brick wall with varied}{earth tones and faded mortar.}{1}\\[.4ex]
        \row{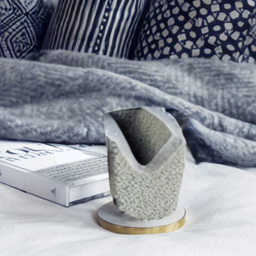}{Grey interlocking roof}{tile texture.}{2}\\[.4ex]
        \row{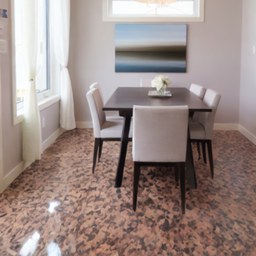}{Dark speckled stone texture}{with irregular black flecks.}{3}\\[.4ex]

        \row{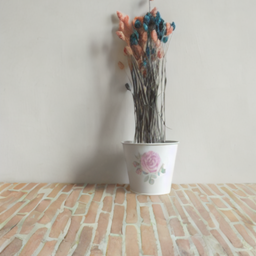}{Red brick wall texture}{with moss growth.}{4}\\[.4ex]
        \row{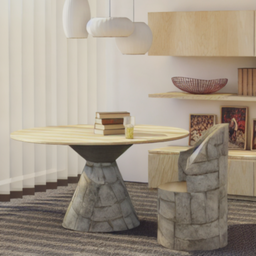}{Grey concrete tile pattern with mossy patches.}{}{5}\\[.4ex]
        \row{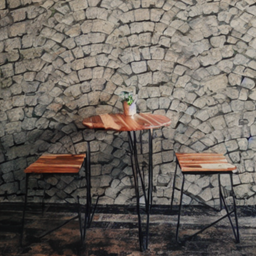}{Grey brick tile texture}{with light mottled specks.}{6}\\[.4ex]
        \row{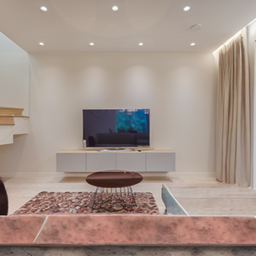}{Weathered reddish-brown}{brick wall texture.}{7}\\[.4ex]

        \row{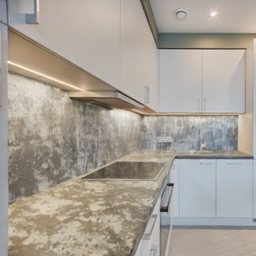}{A distressed gray concrete wall}{texture with white stains.}{8}\\[.4ex]
        \row{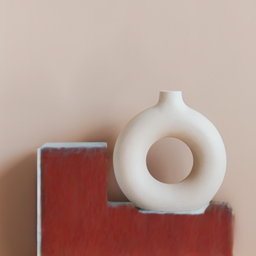}{Red brick texture with visible}{mortar lines and chipped edges.}{9}\\[.4ex]
        \row{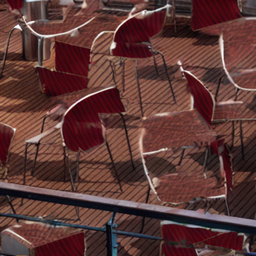}{Dark maroon cloth texture with}{crisscrossed stitching patterns.}{10}
                
    \end{tabular}}
    \caption{\textbf{Comparison to baselines.} We compare against text-based (Blended LD, FLUX) and image-based methods (SD2-ControlNet-Normal, \rgbtooxx, ZeST, Material Fusion). Each method uses the input image and mask information either via latent masking or as explicit conditioning.
    Our approach better preserves the target image lighting while maintaining the transferred material appearance.}
    \label{fig:baselines}
\end{figure*}

\begin{figure}[hbt!]
    \centering
    \setlength{\tabcolsep}{0pt}
    \renewcommand{\arraystretch}{0}
    \newcommand{\imgsize}{.25\linewidth}
    \newcommand{\imgsizehalf}{.125\linewidth}

    \newcommand{\row}[1]{%
        &\multicolumn{2}{c}{\includegraphics[width=\imgsize,height=\imgsize]{images/artefacts/texture_#1.png}}%
        && \multicolumn{2}{c}{\includegraphics[width=\imgsize,height=\imgsize]{images/artefacts/zest_#1.png}}%
        && \multicolumn{2}{c}{\includegraphics[width=\imgsize,height=\imgsize]{images/artefacts/matfusion_#1.png}}%
        && \multicolumn{2}{c}{\includegraphics[width=\imgsize,height=\imgsize]{images/artefacts/ours_#1.png}} \\%
        
        &\includegraphics[width=\imgsizehalf,height=\imgsizehalf]{images/artefacts/image_#1.png}%
        &\includegraphics[width=\imgsizehalf,height=\imgsizehalf]{images/artefacts/overlay_#1.png}%
        &&\includegraphics[width=\imgsizehalf,height=\imgsizehalf]{images/artefacts/zest_#1_crop1.png}%
        & \includegraphics[width=\imgsizehalf,height=\imgsizehalf]{images/artefacts/zest_#1_crop2.png}%
        && \includegraphics[width=\imgsizehalf,height=\imgsizehalf]{images/artefacts/matfusion_#1_crop1.png}%
        & \includegraphics[width=\imgsizehalf,height=\imgsizehalf]{images/artefacts/matfusion_#1_crop2.png}%
        && \includegraphics[width=\imgsizehalf,height=\imgsizehalf]{images/artefacts/ours_#1_crop1.png}%
        & \includegraphics[width=\imgsizehalf,height=\imgsizehalf]{images/artefacts/ours_#1_crop2.png}\\[0.5ex]
    }
    
	\resizebox{.99\linewidth}{!}{%
    \begin{tabular}{c cc C{0.5ex} cc C{0.3ex} cc C{0.3ex} cc}
        &\multicolumn{2}{c}{Conditions} && \multicolumn{2}{c}{ZeST} && \multicolumn{2}{c}{Material Fusion} && \multicolumn{2}{c}{Ours} \\[0.5ex]

        \row{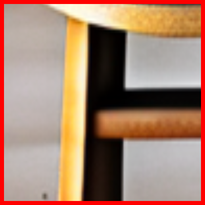}\\
        \row{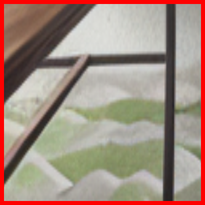}
    \end{tabular}}
    \caption{\textbf{Limitations of prior methods.} We illustrate some of the limitations of recent material transfer methods. Both ZeST and Material Fusion can affect more than the region defined by the mask and can also severely impact the object's geometry. ZeST has some border effects because of its latent masking and entangled semantic information (\cf stool leg turned into a chess piece). Thanks to our training, the material identity and object geometry are preserved.}
    \label{fig:artefacts}
\end{figure}

\section{Experiments}

We now compare our method against state-of-the-art inpainting and material transfer methods.

\textbf{Baselines.} We consider two types of inpainting baselines. 
First, we look at specialized material editing baselines using visual-prompt guidance including ZeST~\cite{cheng2024zest}, MaterialFusion~\cite{garifullin2025materialfusion}, and \rgbtooxx~\cite{zeng2024rgbx}. We further include a variant of ZeST -- ``SD2 Ctrl-\normal'' -- based on the normals-conditioned ControlNet~\cite{zhang2023controlnet} of the SD v2.1 model. Since \rgbtooxx operates in camera perspective, the method is not perfectly suitable for our task without UV mapping. Nevertheless, we first decompose the image (\rgbtoxx), project the target texture onto the \xx-albedo map, and execute \xxtorgb to recompose the image. 
Moreover, we extend our comparisons to text-prompt inpainting methods. We cover the following publicly available models: Stable Diffusion v2.1~\cite{rombach2021highresolution}, the inpainting SD-XL model~\cite{podell2023sdxl}, and the FLUX.1 inpainting model~\cite{flux.1}. Lastly, the recent Blended Latent Diffusion method~\cite{avrahami2023blended} was added. We rely on InternVL2-8B~\cite{chen2024internvl} to caption the textures with short descriptions.
In all baselines, we employ the checkpoints or code provided by the authors.

\textbf{Data.} We conduct our quantitative analysis on 300 pairs of synthetic renders (\cf~\cref{fig:archviz}) and 50 real images (Figs.~\ref{fig:baselines},~\ref{fig:artefacts},~\ref{fig:irradiance}). The synthetic test set includes artist-made 3D scenes~\cite{evermotion} rendered with Blender's physically-based Cycles renderer~\cite{blender}. We render the images and the ground truth normals and irradiance maps. Unless stated otherwise, the images shown in the paper are photographs sourced from the Materialistic evaluation set \cite{sharma2023materialistic}. All our evaluation data will be publicly released. 

\textbf{Metrics.} We evaluate synthetic data using PSNR and LPIPS \cite{zhang2018unreasonable}. Given that real data lacks ground truth maps, we evaluate its appearance using CLIP-I~\cite{radford2021learning} by computing the cosine similarity score between the exemplar image and a crop of the generated region. Additionally, we compare its estimated irradiance to that of the original target image to assess the adherence to lighting cues. We report PSNR and LPIPS over the target region.

\subsection{Main results}

\begin{table}
	\centering
	\scriptsize
    \setlength{\tabcolsep}{3pt}
    
    \newcommand{\tick}{\ding{51}}
    
	\resizebox{.87\linewidth}{!}{%
    \begin{tabular}{c | ccH | c}
        \toprule
        \multirow{2}{*}{Method} & \multicolumn{3}{c|}{Synthetic} & Real \\
        & PSNR$_\uparrow$ & LPIPS$_\downarrow$ & SSIM$_\uparrow$ & CLIP-I$_\uparrow$ \\
        \midrule
        SD v2.1 \cite{rombach2021highresolution} & 18.16 & 0.2214 &  & 0.7611 \\
        SD-XL inpaint \cite{podell2023sdxl} & 19.10 & 0.2025 & & 0.7462 \\
        Blended LD \cite{avrahami2023blended} & 19.57 & 0.2282 &  & 0.7303 \\
        FLUX.1 Fill \cite{flux.1} & \scnd{19.92} & \scnd{0.1825} &  & 0.7552 \\
        \midrule
        SD2 Ctrl-\normal \cite{zhang2023controlnet} & 18.16 & 0.2142 & & 0.7701 \\
        \rgbtooxx \cite{zeng2024rgbx} & 11.20 & 0.4130 && \scnd{0.7900} \\
        ZeST \cite{cheng2024zest} & 19.10 & 0.1879 & & 0.7790 \\
        Material Fusion \cite{garifullin2025materialfusion} & 19.43 & 0.1916 & & 0.7560 \\
        \oc{}\method{}~(ours) & \oc{}\best{20.62} & \oc{}\best{0.1783} & \oc{} & \oc{}\best{0.7994} \\

        \toprule
    \end{tabular}}
    
	\caption{\textbf{Inpainting results.} Comparison with inpainting baselines~(top) and other material transfer methods (bottom) on synthetic and real scenes for material transfer. 
    }
    \vspace{-4mm}

	\label{tab:generation}
\end{table}
Quantitative results are reported in \cref{tab:generation}. For this experiment, all conditionings are provided to the methods. While inpainting methods (top four rows) demonstrate competitive performance, they do not match the effectiveness of specialized methods for material transfer (bottom five rows). On synthetic data, FLUX.1 shows competitive performance, even beating ZeST, which specializes in material transfer. On this data, our method shows improvements of $+3.5\%$ in PSNR and $+2.3\%$ in LPIPS compared to the second-best performing method. On real data, we achieve a $+1.2\%$ improvement on the CLIP-I measure over \rgbtooxx{}, the next best-performing method. Overall, our method establishes a new state-of-the-art on all evaluated metrics.

We further evaluate the shading error produced by each method in \cref{tab:irradiance}. This error is determined by comparing the estimated irradiance maps of the output image $\phi_\irra(\hat{\image})$ and the reference image $\phi_\irra(\image)$. We observe that newer inpainting methods based on FLUX.1 and Blended LD preserve the illumination from the original image well. Our method, guided by the irradiance map \irra{}, understandably outperforms all compared methods in illumination preservation. 

We present qualitative results in \cref{fig:baselines}. We note that earlier inpainting methods such as Stable Diffusion based methods (SD2 Ctrl-N \cite{zhang2023controlnet}, \rgbtooxx{} \cite{zeng2024rgbx}) have trouble with perspective projection, often offering an orthographic view of the material pasted directly into the region (\edit{rows 2, 5, and 8}), greatly hindering the realism of the edits. Newer methods such as FLUX.1 \cite{flux.1} and Blended LD \cite{avrahami2023blended} better adhere to the scene's geometry, but either lack perspective for FLUX.1 (seventh row) or differ from the exemplar material \exemplar{} (\edit{rows 4 and 9}). ZeST generally provides good geometry coherence, but exhibits artifacts (\edit{rows 2, 4, and 7}). MaterialFusion \cite{garifullin2025materialfusion} often fixes these perspective issues, but often fails to transfer the correct material (\edit{rows 2, 4, and 10}). In addition to good perspective projection (\edit{row 6}) and good adherence to the exemplar material \exemplar{} (\edit{row 4}), our method \method{} provides more complex lighting interactions as reflections and highlights from lights (\edit{rows 1 and 3}). In general, \method{} produces material transfers that blend well with their surroundings while preserving illumination on the applied material. 
\begin{figure}
    \centering
    \setlength{\tabcolsep}{0pt}
    \renewcommand{\arraystretch}{0}
    \newcommand{\imgsize}{.2\linewidth}
    \newcommand{\imgsizehalf}{.1\linewidth}
    \newcommand{\shift}{5.9ex}
    
    \newcommand{\row}[2]{%
        &\includegraphics[width=\imgsizehalf,height=\imgsizehalf]{images/complex/irradiance_#1.png}%
        & \includegraphics[width=\imgsizehalf,height=\imgsizehalf]{images/complex/image_#1.png}%
        & \multirow{2}{*}[\shift]{\includegraphics[width=\imgsize,height=\imgsize]{images/complex/texture_#1.png}}%
        & \multirow{2}{*}[\shift]{\includegraphics[width=\imgsize,height=\imgsize]{images/complex/ours_#1_cfg03.png}}%
        & \includegraphics[width=\imgsizehalf,height=\imgsizehalf]{images/complex/irradiance_#2.png}%
        & \includegraphics[width=\imgsizehalf,height=\imgsizehalf]{images/complex/image_#2.png}%
        & \multirow{2}{*}[\shift]{\includegraphics[width=\imgsize,height=\imgsize]{images/complex/texture_#2.png}}%
        & \multirow{2}{*}[\shift]{\includegraphics[width=\imgsize,height=\imgsize]{images/complex/ours_#2_cfg03.png}} \\
        
        &\includegraphics[width=\imgsizehalf,height=\imgsizehalf]{images/complex/normals_#1.png}%
        & \includegraphics[width=\imgsizehalf,height=\imgsizehalf]{images/complex/mask_#1.png}%
        &&& \includegraphics[width=\imgsizehalf,height=\imgsizehalf]{images/complex/normals_#2.png}%
        & \includegraphics[width=\imgsizehalf,height=\imgsizehalf]{images/complex/mask_#2.png} \\[.3ex]
    }
    
	\resizebox{.99\linewidth}{!}{%
    \begin{tabular}{c Hccc S{.5ex} Hccc}
        &\multicolumn{2}{c}{} & Texture & Output & \multicolumn{2}{c}{} & Texture & Output \\[.7ex]
        
        \row{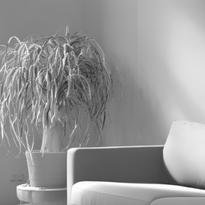}{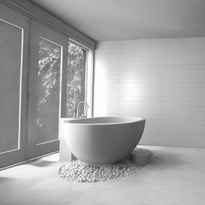} 
        \row{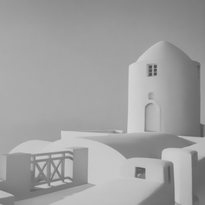}{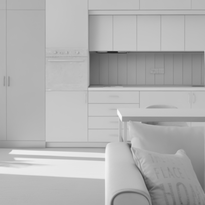} %
        \row{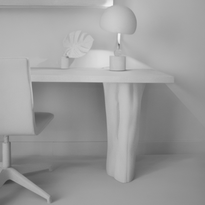}{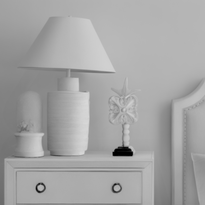}
        
    \end{tabular}}
    \caption{\edit{\textbf{Non-planar surfaces.} We provide results on non-planar surfaces, demonstrating that our method is capable of handling surfaces with more complex geometry. Zoom in for details.}}
    \label{fig:complex}
    \vspace{-6mm}
\end{figure}

Considering \method{} takes normals map with both the geometric and material level variation as input, it is able to disentangle them and ignore the original material normals in the generated image \target. In \cref{fig:baselines}, \edit{row 3}, \method{} ignores the wooden planks normals while respecting the floor general geometry %
resulting in a flat surface (black and pink marble). In contrast, ZeST retains these cues (notice lines across the floor), despite mismatched semantics of the newly transferred material. Similarly, our method adapts well to non-flat surfaces such as in \edit{row 2}, where the applied material fits the curved shape of the mug. The official implementation of \cite{cheng2024zest} uses the original SDXL checkpoint which is not finetuned for inpainting. This means that the method suffers from noticeable artifacts around the target region, zoom in on \cref{fig:baselines}.
We further evaluate limitations of recent transfer methods in \cref{fig:artefacts}, highlighting masking issues. 

\begin{figure*}
    \centering
	\large
    \setlength{\tabcolsep}{0pt}
    \renewcommand{\arraystretch}{0}
    \newcommand{\imgsize}{.125\textwidth}
    \newcommand{\imgsizehalf}{.0625\textwidth}

    \newcommand{\row}[3]{%
        \multirow{2}{*}[5ex]{\verti{0ex}{\edit{row #3}}}
        & \multirow{2}{*}[#2]{\includegraphics[width=\imgsize]{images/light/image_#1.png}}%
        & \multirow{2}{*}[#2]{\includegraphics[width=\imgsize]{images/light/irradiance_#1.png}}%
        & \includegraphics[width=\imgsizehalf]{images/light/texture_#1.png}%
        & \multirow{2}{*}[#2]{\includegraphics[width=\imgsize]{images/light/zest_#1.png}}%
        & \multirow{2}{*}[#2]{\includegraphics[width=\imgsize]{images/light/zest_#1_irradiance.png}}%
        & \multirow{2}{*}[#2]{\includegraphics[width=\imgsize]{images/light/mocka_v3_a100skip_nomaskp_A5_#1.png}}%
        & \multirow{2}{*}[#2]{\includegraphics[width=\imgsize]{images/light/mocka_v3_a100skip_nomaskp_A5_#1_irradiance.png}}%
        
        & \multirow{2}{*}[#2]{\includegraphics[width=\imgsize]{images/light/mocka_v3_a100_512px_#1.png}}%
        & \multirow{2}{*}[#2]{\includegraphics[width=\imgsize]{images/light/mocka_v3_a100_512px_#1_irradiance.png}} \\%
        
        &&&\includegraphics[width=\imgsizehalf]{images/light/overlay_#1.png} \\
    }
    
	\resizebox{.99\linewidth}{!}{%
    \begin{tabular}{C{2ex} cc S{.5ex} c S{.5ex} cc S{1ex} cc S{1ex} cc}
        & \multicolumn{2}{c}{Image \& Irradiance} & Source & \multicolumn{2}{c}{ZeST} & \multicolumn{2}{c}{ours w/o \irra} & \multicolumn{2}{c}{\textbf{ours}} \\[.7ex]

        \row{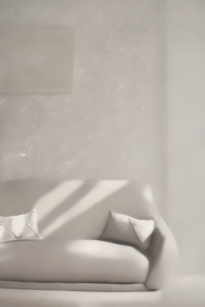}{10.5ex}{1}\\[.2ex] 
        \row{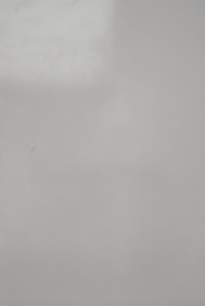}{10.5ex}{2}\\[.2ex] 
        \row{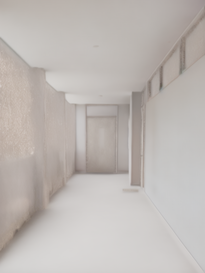}{9.3ex}{3}\\[.2ex]
        \row{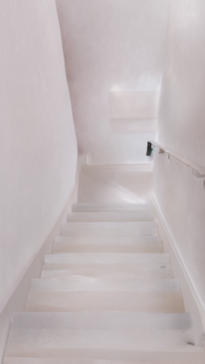}{12.4ex}{4}

        \cmidrule[1pt](lr){2-3} \cmidrule[1pt](lr){5-6} \cmidrule[1pt](lr){7-8} \cmidrule[1pt](lr){9-10}
        & \target & $\phi_\irra(\target)$ & & $\hat{\image}$ & $\phi_\irra(\hat{\image})$ & $\hat{\image}$ & $\phi_\irra(\hat{\image})$ & $\hat{\image}$ & $\phi_\irra(\hat{\image})$ \\
    \end{tabular}}
    \caption{\textbf{Adherence to irradiance.} We compare the irradiance of the input image \target directly against the irradiance estimated from the images $\hat{\image}$ edited by ZeST and our model (\edit{with or without training using the irradiance map \irra}). Column ``ours w/o \irra'' corresponds to ablation $(A_5)$ from \cref{tab:ablation}. Compared to our primary baseline, our model receives information from both the irradiance and the non-masked image, which allows us to better preserve the illumination of the original scene. \edit{This is clearly seen with light properly illuminating the wall on the first row, and the claret-colored wall reflecting light at a grazing angle on the last row.}
    }
    \label{fig:irradiance}
    \vspace{-3mm}
\end{figure*}
\begin{table}
	\centering
	\scriptsize
	\resizebox{.75\linewidth}{!}{%
		\begin{tabular}{c | cc}
			\toprule
			Method & PSNR$_\uparrow$ & LPIPS$_\downarrow$ \\
			\midrule
            SD v2.1 \cite{rombach2021highresolution} & 16.79 & 0.1275 \\
            SD-XL inpaint \cite{podell2023sdxl} & 18.86 & 0.1010 \\
            Blended LD \cite{avrahami2023blended} & 20.40 & 0.0847 \\
            FLUX.1 Fill \cite{flux.1} & \scnd{20.93} & 0.0821 \\
            \midrule
            SD2 Ctrl-\normal \cite{zhang2023controlnet} & 17.01 & 0.1136 \\
            \rgbtooxx \cite{zeng2024rgbx} & 20.76 & \scnd{0.0730} \\
            ZeST \cite{cheng2024zest} & 18.84 & 0.0962 \\
            Material Fusion \cite{garifullin2025materialfusion} & 20.01 & 0.0788 \\
            \method{}~w/o \irra & 18.94 & 0.0925 \\
            \oc{}\method{}~(ours) & \oc{}\best{21.43} & \oc{}\best{0.0668}  \\

			\toprule
	\end{tabular}}
	\caption{\textbf{Adherence to irradiance.} We measure the shading error by estimating the irradiance map of the model output, \ie, $\phi_\irra(\hat{\image})$. We then compute its quality against the irradiance of the original image, \ie, $\phi_\irra(x)$. This evaluates the adherence of the model to the illumination existing in the original image.}
	\label{tab:irradiance}
    \vspace{-4mm}
\end{table}

\begin{figure*}
    \centering

    \setlength{\tabcolsep}{0pt}
    \renewcommand{\arraystretch}{0}
    \newcommand{\imgsize}{0.143\textwidth}
    \newcommand{\imgsizehalf}{.071\textwidth}

    \newcommand{\imgsizeh}{0.2\textwidth}
    \newcommand{\imgsizehalfh}{.1\textwidth}

    \newcommand{\rowvertical}[7]{%
        &\includegraphics[width=\imgsizehalf]{images/hue/irradiance_#1.png}%
        & \multirow{2}{*}[#2]{\includegraphics[width=\imgsize]{images/hue/image_#1.png}}%
        & \includegraphics[width=\imgsizehalf]{images/hue/texture_#1_#3.png}%
        && \multirow{2}{*}[#2]{\includegraphics[width=\imgsize]{images/hue/ours_#1_#3.png}}%
        & \multirow{2}{*}[#2]{\includegraphics[width=\imgsize]{images/hue/ours_#1_#4.png}}%
        & \multirow{2}{*}[#2]{\includegraphics[width=\imgsize]{images/hue/ours_#1_#5.png}}%
        & \multirow{2}{*}[#2]{\includegraphics[width=\imgsize]{images/hue/ours_#1_#6.png}}%
        & \multirow{2}{*}[#2]{\includegraphics[width=\imgsize]{images/hue/ours_#1_#7.png}}%
        \\
        
        & \includegraphics[width=\imgsizehalf]{images/hue/normals_#1.png}%
        && \includegraphics[width=\imgsizehalf]{images/hue/mask_#1.png}%
        \\
    }

    \newcommand{\rowhorizontal}[5]{%
        &\includegraphics[width=\imgsizehalfh]{images/hue/irradiance_#1.png}%
        & \multirow{2}{*}[#2]{\includegraphics[width=\imgsizeh]{images/hue/image_#1.png}}%
        & \includegraphics[width=\imgsizehalfh]{images/hue/texture_#1_#3.png}%
        && \multirow{2}{*}[#2]{\includegraphics[width=\imgsizeh]{images/hue/ours_#1_#3.png}}%
        & \multirow{2}{*}[#2]{\includegraphics[width=\imgsizeh]{images/hue/ours_#1_#4.png}}%
        & \multirow{2}{*}[#2]{\includegraphics[width=\imgsizeh]{images/hue/ours_#1_#5.png}}%
        \\
        
        & \includegraphics[width=\imgsizehalfh]{images/hue/normals_#1.png}%
        && \includegraphics[width=\imgsizehalfh]{images/hue/mask_#1.png}%
        \\
    }

    \newcommand{\cbull}[2]{\textcolor[HTML]{#2}{\raisebox{-0ex}{\scalebox{#1}{$\bullet$}}}}
    
	\resizebox{.99\linewidth}{!}{%
    \begin{tabular}{c ccS{0.5ex}c C{0.5ex} ccccc}

        &\multicolumn{2}{c}{Conditions} & Source &&\cbull{1.5}{804916} & \cbull{1.5}{388016} & \cbull{1.5}{1f1680} & \cbull{1.5}{5e1680} & \cbull{1.5}{801623} \\[0.5ex]
        \rowvertical{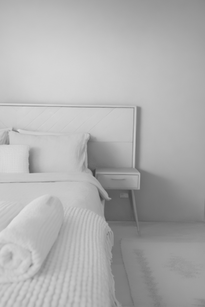}{13.4ex}{h000_804916}{h036_388016}{h108_1f1680}{h126_5e1680}{h162_801623}\\[0.5ex]

        && &  && \cbull{1.5}{b8bf9e} & \cbull{1.5}{9dc0ab} & \cbull{1.5}{9eabc0} & \cbull{1.5}{b89ebf} & \cbull{1.5}{c0a09f} \\[0.5ex]
        \rowvertical{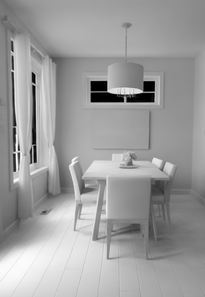}{12.8ex}{h018_b8bf9e}{h054_9dc0ab}{h090_9eabc0}{h126_b89ebf}{h162_c0a09f} \\[3ex]
    \end{tabular}}
    
    \newcommand{\cbullo}[2]{\textcolor[HTML]{#2}{\raisebox{-0ex}{\scalebox{#1}{$\bullet$}}}}
    
    \resizebox{.99\linewidth}{!}{%
    \begin{tabular}{c ccS{0.5ex}c C{0.5ex} ccc}
        &\multicolumn{2}{c}{Conditions} & Source && \cbullo{1.5}{aa9982} & \cbullo{1.5}{aa8383} & \cbullo{1.5}{a182a9} \\[0.5ex]
        \rowhorizontal{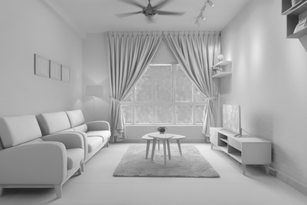}{8.4ex}{h000_aa9982}{h162_aa8383}{h126_a182a9}\\[1ex]

        &&&&& \cbullo{1.5}{c0bed8} & \cbullo{1.5}{c7d8be} & \cbullo{1.5}{d8bec2} \\[0.5ex]
        \rowhorizontal{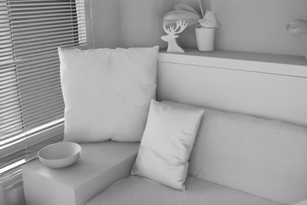}{8.5ex}{h126_c0bed8}{h054_c7d8be}{h000_d8bec2}\\[1ex]
        
        &&&&& \cbullo{1.5}{9bb9bd} & \cbullo{1.5}{bdb39b} & \cbullo{1.5}{b99bbd} \\[0.5ex]
        \rowhorizontal{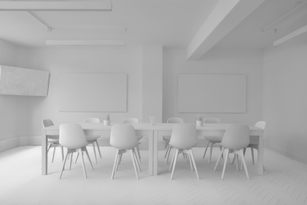}{8.5ex}{h072_9bb9bd}{h000_bdb39b}{h126_b99bbd}\\[1ex]
        
        &&&&& \cbullo{1.5}{4e373b} & \cbullo{1.5}{37374e} & \cbullo{1.5}{374e3c} \\[0.5ex]
        \rowhorizontal{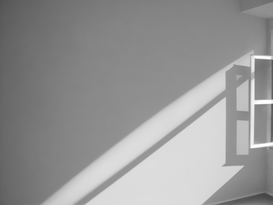}{9.3ex}{h162_4e373b}{h108_37374e}{h054_374e3c}
    \end{tabular}}
    
    \caption{\textbf{Adherence to texture conditioning.} We provide different hue variations of the exemplar material as input and observe that our method correctly adapts to it, maintaining realism in the generated image. \edit{This shows the robustness of our texture conditioning approach.}}
    \label{fig:variations}
\end{figure*}

\begin{figure}
    \centering
    \setlength{\tabcolsep}{0pt}
    \renewcommand{\arraystretch}{0}
    \newcommand{\imgsize}{.2\linewidth}
    \newcommand{\imgsizehalf}{.1\linewidth}
    \newcommand{\shift}{5.9ex}

    \newcommand{\cbullo}[2]{\textcolor[HTML]{#2}{\raisebox{-0ex}{\scalebox{#1}{$\bullet$}}}}
    
    \newcommand{\row}[1]{%
        & \includegraphics[width=\imgsizehalf,height=\imgsizehalf]{images/scale/irradiance_#1.png}%
        & \includegraphics[width=\imgsizehalf,height=\imgsizehalf]{images/scale/image_#1.png}%
        & \multirow{2}{*}[\shift]{\includegraphics[width=\imgsize,height=\imgsize]{images/scale/texture_#1.png}}%
        & \multirow{2}{*}[\shift]{\includegraphics[width=\imgsize,height=\imgsize]{images/scale/ours_#1_z0.png}}%
        & \multirow{2}{*}[\shift]{\includegraphics[width=\imgsize,height=\imgsize]{images/scale/ours_#1_z1.png}}%
        & \multirow{2}{*}[\shift]{\includegraphics[width=\imgsize,height=\imgsize]{images/scale/ours_#1_z2.png}} \\%
        
        &\includegraphics[width=\imgsizehalf,height=\imgsizehalf]{images/scale/normals_#1.png} %
        & \includegraphics[width=\imgsizehalf,height=\imgsizehalf]{images/scale/overlay_#1.png} \\[0.4ex]%
    }
    
	\resizebox{.99\linewidth}{!}{%
    \begin{tabular}{c Hc S{.2ex} c S{.2ex} ccc}
        &\multicolumn{3}{c}{Conditions} & $\times 1$ \cbullo{1.5}{00FFFF} & $\times 2$ \cbullo{1.5}{FD3DB5} & $\times 4$ \cbullo{1.5}{FFDE21} \\[.7ex]

        \row{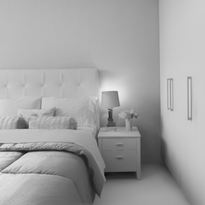}
        \row{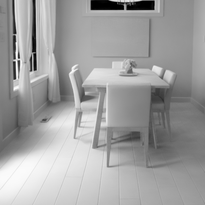}
        \row{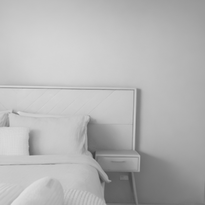}
        \row{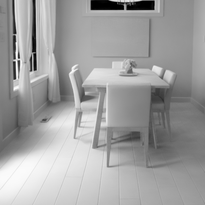}
        \row{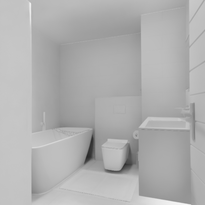}
    \end{tabular}}
    \caption{\textbf{Impact of exemplar scale.} We can control the scale of the transfered material by cropping \exemplar{}. We show results using the full material ($\times{}\!1$), half-sized crop ($\times{}\!2$), and a quarter-sized crop ($\times{}\!4$) to observe its effect on resulting image. Our model has learned to properly interpret this characteristic from the CLIP features.
    }
    \label{fig:scale}
\end{figure}
\begin{figure}
    \centering
    \setlength{\tabcolsep}{1pt}
    \renewcommand{\arraystretch}{0}
    \newcommand{\imgsize}{.10\textwidth}
    \newcommand{\imgsizehalf}{.05\textwidth}

    \newcommand{\row}[2]{%
        & \includegraphics[width=\imgsizehalf]{images/dropout/texture_#1.png}%
        & \multirow{2}{*}[#2]{\includegraphics[width=\imgsize]{images/dropout/ours_#1_drop_N_masked.png}}%
        & \multirow{2}{*}[#2]{\includegraphics[width=\imgsize]{images/dropout/ours_#1_drop_E_masked.png}}%
        & \multirow{2}{*}[#2]{\includegraphics[width=\imgsize]{images/dropout/ours_#1_drop_NE_masked.png}}%
        & \multirow{2}{*}[#2]{\includegraphics[width=\imgsize]{images/dropout/ours_#1_masked.png}}%
        & \multirow{2}{*}[#2]{\includegraphics[width=\imgsize]{images/dropout/ours_#1_drop_N.png}}%
        & \multirow{2}{*}[#2]{\includegraphics[width=\imgsize]{images/dropout/ours_#1_drop_E.png}}%
        & \multirow{2}{*}[#2]{\includegraphics[width=\imgsize]{images/dropout/ours_#1_drop_NE.png}}%
        & \multirow{2}{*}[#2]{\includegraphics[width=\imgsize]{images/dropout/ours_#1.png}}\\
        
        & \includegraphics[width=\imgsizehalf]{images/dropout/mask_#1.png} &&&& \\[0.1ex]
    }
    
	\resizebox{.99\linewidth}{!}{%
    \begin{tabular}{c cS{0.2ex} HcHc HcHc}
        & & \multicolumn{4}{c}{target: $\target\cdot (1\text{-}\mask)$} & \multicolumn{4}{c}{target: $\target$}\\
        \cmidrule[1pt](lr){3-6} \cmidrule[1pt](lr){7-10}
        & Src. & $\xx \setminus \{ \normal \}$ & $\xx \setminus \{ \irra \}$ & $\xx \setminus \{ \normal,\irra \}$ & $\xx$ & $\xx \setminus \{ \normal \}$ & $\xx \setminus \{ \irra \}$ & $\xx \setminus \{ \normal,\irra \}$ & $\xx$ \\[.7ex]
        
        \row{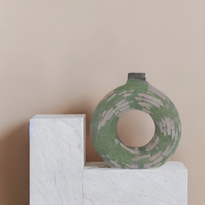}{6.2ex}\\
        \row{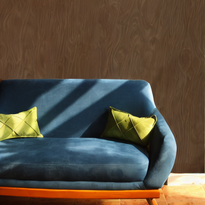}{6.3ex}\\[.2ex]
        
    \end{tabular}}
    
    \caption{\textbf{Ablation on lighting cues.} When deprived of lighting cues by masking out the target image \target{} (\ie, providing $\target\!\cdot\!(1\text{-}\mask)$ as target) and removing the irradiance map \irra{}, our method produces results with flat, implausible shading (leftmost). Reintroducing either the irradiance \irra{} (second column) or the masked region (third column) restores the light effects. Providing all lighting cues (\method{}) provides the best result (rightmost).
    }
    \label{fig:dropout}
\end{figure}

We evaluate the importance of the irradiance map~\irra{} in~\cref{fig:irradiance}. Our approach generates better matching shading than previous work, even without the irradiance map. However, high-frequency lighting effects from the original image such as highlights (first to third rows) are better preserved with the irradiance map.\\
\indent\edit{We present qualitative results on non-planar surfaces in~\cref{fig:complex}. \method{} produces plausible projections on objects with non-planar geometries, such as the table leg and the stone couch. This demonstrates the capabilities of \method{} to transfer materials beyond simple flat surfaces.} \\
\indent Additional analysis on color control can be found in~\cref{fig:variations}. For these results, we convert the exemplar material \exemplar{} from RGB to HSV and change its hue. Our method respects the user-defined color well \edit{even when the specified hues were not explicitly present in the training set}.\\
\indent Finally, we demonstrate our ability to control the scale of the inpainted material by adjusting the scale of the exemplar~\exemplar{}. We evaluate this effect in~\cref{fig:scale} with three zoom levels. As our method processes larger features, it scales them up in the scene accordingly. 

\subsection{User study}
We conducted a two-alternative forced choice (2AFC) study consisting of 74 questions, on a total of 40 participants. The aim is to judge both the realism and fidelity of our method compared to our most competitive baselines:  ZeST \cite{cheng2024zest}, MaterialFusion \cite{garifullin2025materialfusion}, and RGBX \cite{zeng2024rgbx}. Our method is judged more realistic 78\% of the time (being 67\%, 74\%, 91\% per method, respectively) and 70\% more reliable in terms of fidelity to the texture condition (48\%, 86\%, 79\% per method, respectively). According to our study, ZeST and our method show similar fidelity, but our results are more realistic two-thirds of the time.

\subsection{Ablation study}
We quantitatively evaluate the impact of each component of our method in \cref{tab:ablation}. All ablations are trained on the entirety of our \datasetname{} dataset, and evaluated on synthetic scenes with CFG disabled. We begin our ablation by training either the IP-Adapter $(A_1)$ or the denoising UNet model $(A_2)$ separately to establish a baseline performance. We also train both in $(A_3)$, which is akin to a fine-tuned version of ZeST with Stable Diffusion v1.5, yielding results comparable to both of the previous baselines. Unfreezing the UNet $(A_2)$ gives freedom for the image and mask to be used as conditionings $(A_{4-8})$, which significantly boosts performance. Introducing the irradiance map~\irra{} $(A_6)$ helps preserve the image shading, thus improving the results. Adding normals~\normal{} $(A_5)$ improves the results slightly; we hypothesize its role is to disambiguate possible confusion between the geometry and the lighting. Overall, training the IP-Adapter slightly improves the result compared to solely fine-tuning the denoising U-Net, keeping the IP-Adapter layers frozen with pretrained weights from~\cite{ye2023ipadapter}. 

\begin{table}[h]
	\centering
    \setlength{\tabcolsep}{2pt}
    \newcommand{\tick}{\ding{51}}
    
	\resizebox{.7\linewidth}{!}{%
		\begin{tabular}{c C{5ex}C{5ex} | C{3ex}C{3ex}C{3ex} | cc}
			\bottomrule
            &\multicolumn{2}{c|}{Train} & \multicolumn{3}{c|}{Maps} & \multicolumn{2}{c}{Metrics} \\
			&IP-A & UNet & \mask & \normal & \irra & PSNR$_\uparrow$ & LPIPS$_\downarrow$ \\
            \midrule
            \multicolumn{1}{c}{$(A_1)$} & \tick &&&&& 19.06 & 0.3560 \\
            \multicolumn{1}{c}{$(A_2)$} & & \tick &&&& 18.81 & 0.3626 \\
            \midrule
            \multicolumn{1}{c}{$(A_3)$} & \tick & \tick &&&& 18.70 & 0.3642 \\
            \multicolumn{1}{c}{$(A_4)$} & \tick & \tick & \tick &&& 20.04 & 0.3277 \\
            \multicolumn{1}{c}{$(A_5)$} & \tick & \tick & \tick & \tick && 19.91 & 0.3284 \\
            \multicolumn{1}{c}{$(A_6)$} & \tick & \tick & \tick && \tick & \scnd{20.28} & 0.3281 \\
            \midrule
            \multicolumn{1}{c}{$(A_7)$} & & \tick & \tick & \tick & \tick & 20.22 & \best{0.3243} \\
            \multicolumn{1}{c}{\oc{}$(A_8)$} &\oc{}\tick & \oc{}\tick & \oc{}\tick & \oc{}\tick & \oc{}\tick & \oc{}\best{20.38} & \oc{}\scnd{0.3256} \\
            
            \bottomrule
	\end{tabular}}
	\caption{\textbf{Ablation study.} We evaluate our main architectural and design decisions. All ablations are trained on the full set for 50k iterations. Our approach significantly benefits from adding the irradiance information $(A_6)$.
    Instead of applying a masked loss as in our baselines  $(A_{1-3})$, we pass the mask as an input to the UNet, which significantly enhances the quality of the material transfer. 
    }
	\label{tab:ablation}
\end{table}

We explore the role of lighting conditioning in \cref{fig:dropout}, showing that our model considers cues from both the target \target{} and the irradiance map \irra{}. As expected, removing all lighting cues significantly deteriorates shading quality. We do so by masking the target region \mask{} in the target image \target{} and removing the irradiance map \irra{}, that corresponds to using $\xx = \{\target\cdot(1\text{-}\mask), \normal, \mask\}$. The best results are obtained when both the full target image and its irradiance are provided which validates the effectiveness of our method. 

\begin{figure}
    \centering
    \setlength{\tabcolsep}{0pt}
    \renewcommand{\arraystretch}{0}
    \newcommand{\imgsize}{.166\linewidth}
    \newcommand{\imgsizehalf}{.083\linewidth}
    \newcommand{\shift}{4.9ex}
    
    \newcommand{\row}[1]{%
        &\includegraphics[width=\imgsizehalf,height=\imgsizehalf]{images/cfg/image_#1.png}%
        & \multirow{2}{*}[\shift]{\includegraphics[width=\imgsize,height=\imgsize]{images/cfg/ours_#1_cfg01.png}}%
        & \multirow{2}{*}[\shift]{\includegraphics[width=\imgsize,height=\imgsize]{images/cfg/ours_#1_cfg03.png}}%
        & \multirow{2}{*}[\shift]{\includegraphics[width=\imgsize,height=\imgsize]{images/cfg/ours_#1_cfg05.png}}%
        & \multirow{2}{*}[\shift]{\includegraphics[width=\imgsize,height=\imgsize]{images/cfg/ours_#1_cfg10.png}}%
        & \multirow{2}{*}[\shift]{\includegraphics[width=\imgsize,height=\imgsize]{images/cfg/ours_#1_cfg20.png}} \\
        
        \multirow{2}{*}[9.8ex]{\includegraphics[width=\imgsize,height=\imgsize]{images/cfg/texture_#1.png}} & \includegraphics[width=\imgsizehalf,height=\imgsizehalf]{images/cfg/overlay_#1.png}  \\[0.1ex]
    }

	\resizebox{.99\linewidth}{!}{%
    \begin{tabular}{cc S{.2ex} cccHc}
        \multicolumn{2}{c}{Conditions} & $\gamma=1$ & $3$ & $5$ & $10$ & $20$ \\
        \row{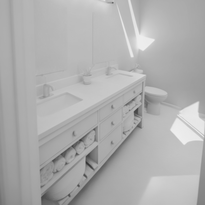}\\[.3ex]
        \row{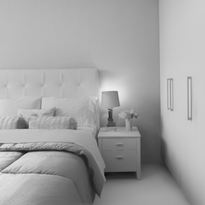}\\[.3ex]
        \row{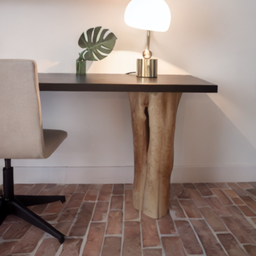}\\[.3ex]
        \row{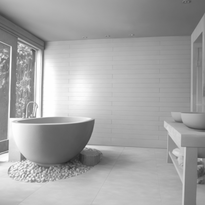}\\
    \end{tabular}}
    \caption{\textbf{Ablation of classifier-free guidance.} We experimentaly found $\gamma=3$ to be a good trade-off between texture realism and fidelity to the conditioning image. Not using CFG leads to misalignments for structured textures as well as artifacts. The $\gamma$ parameter can be changed by the user. Zoom in on the figure for details.}
    \label{fig:cfg}
\end{figure}

Finally, we investigate the impact of the Classifier-Free Guidance \cite{ho2021classifier} in~\cref{fig:cfg}. We notice that using $\gamma \! = \! 3$ improved realism while preserving fidelity compared to setting $\gamma \! = \! 1$. Applying a stronger guidance leads to deteriorations.

\begin{figure}
    \centering
    \setlength{\tabcolsep}{0pt}
    \renewcommand{\arraystretch}{0}
    \newcommand{\imgsize}{.18\linewidth}
    \newcommand{\imgsizehalf}{.09\linewidth}
    \newcommand{\shift}{5.4ex}
    
    \newcommand{\row}[2]{%
        &\includegraphics[width=\imgsizehalf,height=\imgsizehalf]{images/failure/irradiance_#1.png}%
        & \includegraphics[width=\imgsizehalf,height=\imgsizehalf]{images/failure/image_#1.png}%
        & \multirow{2}{*}[\shift]{\includegraphics[width=\imgsize,height=\imgsize]{images/failure/texture_#1.png}}%
        & \multirow{2}{*}[\shift]{\includegraphics[width=\imgsize,height=\imgsize]{images/failure/ours_#1.png}}%
        & \includegraphics[width=\imgsizehalf,height=\imgsizehalf]{images/failure/irradiance_#2.png}%
        & \includegraphics[width=\imgsizehalf,height=\imgsizehalf]{images/failure/image_#2.png}%
        & \multirow{2}{*}[\shift]{\includegraphics[width=\imgsize,height=\imgsize]{images/failure/texture_#2.png}}%
        & \multirow{2}{*}[\shift]{\includegraphics[width=\imgsize,height=\imgsize]{images/failure/ours_#2.png}} \\
        
        &\includegraphics[width=\imgsizehalf,height=\imgsizehalf]{images/failure/normals_#1.png}%
        & \includegraphics[width=\imgsizehalf,height=\imgsizehalf]{images/failure/mask_#1.png}%
        &&& \includegraphics[width=\imgsizehalf,height=\imgsizehalf]{images/failure/normals_#2.png}%
        & \includegraphics[width=\imgsizehalf,height=\imgsizehalf]{images/failure/mask_#2.png} \\[.7ex]
    }
    
	\resizebox{.99\linewidth}{!}{%
    \begin{tabular}{c Hccc S{.5ex} Hccc}
        &\multicolumn{2}{c}{} & Texture & Output & \multicolumn{2}{c}{} & Texture & Output \\[.7ex]
        
        \row{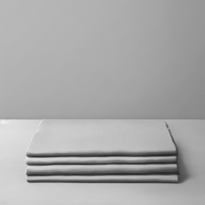}{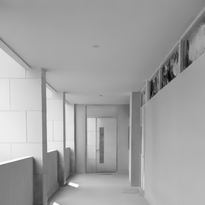}
        \row{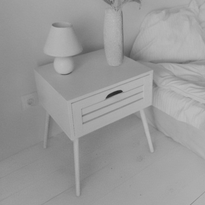}{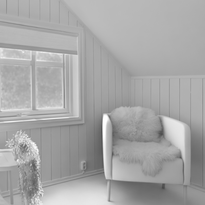}
        
    \end{tabular}}
    \caption{\textbf{Limitations.} We illustrate the limitations of our method, where some geometry can be lost during transfer (left column) and when dealing with downward-facing normals (right column).}
    \label{fig:limitations}
    \vspace{-4mm}
\end{figure}

\section{Limitations}

Despite state-of-the-art performance in material transfer, our method suffers from a few limitations illustrated in~\cref{fig:limitations}. A challenging case is surfaces pointing downwards or exhibiting high-frequency normals, both of which do not appear in our dataset and could benefit from a more extensive training set. Enriching the dataset with additional and more complex objects could improve performance, as done in \cite{sharma2024alchemist}. Another challenge comes from thin or small objects that are difficult to process due to the diffusion model resolution and the downsizing of the input mask. 
Lastly, while \method{} applies albedo changes well on surfaces (see~\cref{fig:variations}) and can produce more glossy or rough surfaces based on its priors -- as shown in \cref{fig:baselines} (row 3) and \cref{fig:irradiance} (row 4) -- it does not provide explicit controls over roughness or placement of the material on the surface.

\section{Conclusion}

We present ~\method{}, a method for material transfer from flat textures to images without requiring complex 3D understanding or UV mapping. Our approach naturally harmonizes the transferred material with the target image illumination, leveraging the available irradiance information. We demonstrate material transfer in real photographs. Our approach provides a more practical tool for artists to edit images and explore possible material variations, for example in the context of architecture visualization and interior design. 

{\small \textbf{Acknowledgments.} This research was funded by the French Agence Nationale de la Recherche (ANR) with project SIGHT, ANR-20-CE23-0016. It was performed using GENCI-IDRIS HPC resources (Grants AD011014389R1, AD011012808R3).}

\printbibliography                

\end{document}